\documentclass[sigconf]{acmart}
\AtBeginDocument{%
  }

\setcopyright{acmlicensed}

\copyrightyear{2025} 
\acmYear{2025} 
\setcopyright{rightsretained}
\acmConference[MMSys '25]{ACM Multimedia Systems Conference 2025}{March 31-April 4, 2025}{Stellenbosch, South Africa}
\acmBooktitle{ACM Multimedia Systems Conference 2025 (MMSys '25), March 31-April 4, 2025, Stellenbosch, South Africa}\acmDOI{10.1145/3712676.3719269}
\acmISBN{979-8-4007-1467-2/2025/03}

\begin{document}

\title{Continual Error Correction on Low-Resource Devices}

\author{Kirill Paramonov}
\authornote{Both authors contributed equally to this research.}
\affiliation{%
  \institution{Samsung R\&D Institute UK}
  \country{United Kingdom}
}
\author{Mete Ozay}
\affiliation{%
  \institution{Samsung R\&D Institute UK}
  \country{United Kingdom}
}

\author{Aristeidis Mystakidis}
\affiliation{%
  \institution{CERTH}
\country{Greece}
}
\author{Nikolaos Tsalikidis}
\affiliation{%
  \institution{CERTH}
\country{Greece}
}
\author{Dimitrios Sotos}
\affiliation{%
  \institution{CERTH}
\country{Greece}
}
\author{Anastasios Drosou}
\affiliation{%
  \institution{CERTH}
\country{Greece}
}
\author{Dimitrios Tzovaras}
\affiliation{%
  \institution{CERTH}
\country{Greece}
}

\author{Hyunjun Kim}
\affiliation{%
  \institution{Samsung Research}
\country{South Korea}
}
\author{Kiseok Chang}
\affiliation{%
  \institution{Samsung Research}
\country{South Korea}
}
\author{Sangdok Mo}
\affiliation{%
  \institution{Samsung Research}
\country{South Korea}
}
\author{Namwoong Kim}
\affiliation{%
  \institution{Samsung Research}
\country{South Korea}
}
\author{Woojong Yoo}
\affiliation{%
  \institution{Samsung Research}
\country{South Korea}
}
\author{Jijoong Moon}
\affiliation{%
  \institution{Samsung Research}
\country{South Korea}
}

\author{Umberto Michieli}
\authornotemark[1]
\authornote{Corresponding author: \texttt{u.michieli@samsung.com}.}
\affiliation{%
  \institution{Samsung R\&D Institute UK}
  \country{United Kingdom}
}

\renewcommand{\shortauthors}{Paramonov et al.}

\newcommand{\um}[1]{\textcolor{red}{UM: #1}}
\newcommand{\mo}[1]{\textcolor{green}{MO: #1}}

\begin{abstract}
The proliferation of AI models in everyday devices has highlighted a critical challenge: prediction errors that degrade user experience. While existing solutions focus on error detection, they rarely provide efficient correction mechanisms, especially for resource-constrained devices. We present a novel system enabling users to correct AI misclassifications through few-shot learning, requiring minimal computational resources and storage. Our approach combines server-side foundation model training with on-device prototype-based classification, enabling efficient error correction through prototype updates rather than model retraining. The system consists of two key components: (1) a server-side pipeline that leverages knowledge distillation to transfer robust feature representations from foundation models to device-compatible architectures, and (2) a device-side mechanism that enables ultra-efficient error correction through prototype adaptation. We demonstrate our system's effectiveness on both image classification and object detection tasks, achieving over 50\% error correction in one-shot scenarios on Food-101 and Flowers-102 datasets while maintaining minimal forgetting (less than 0.02\%) and negligible computational overhead. Our implementation, validated through an Android demonstration app, proves the system's practicality in real-world scenarios.
\end{abstract}



\begin{teaserfigure}
  \includegraphics[width=\textwidth, trim=0cm 13cm 4.5cm 0cm, clip]{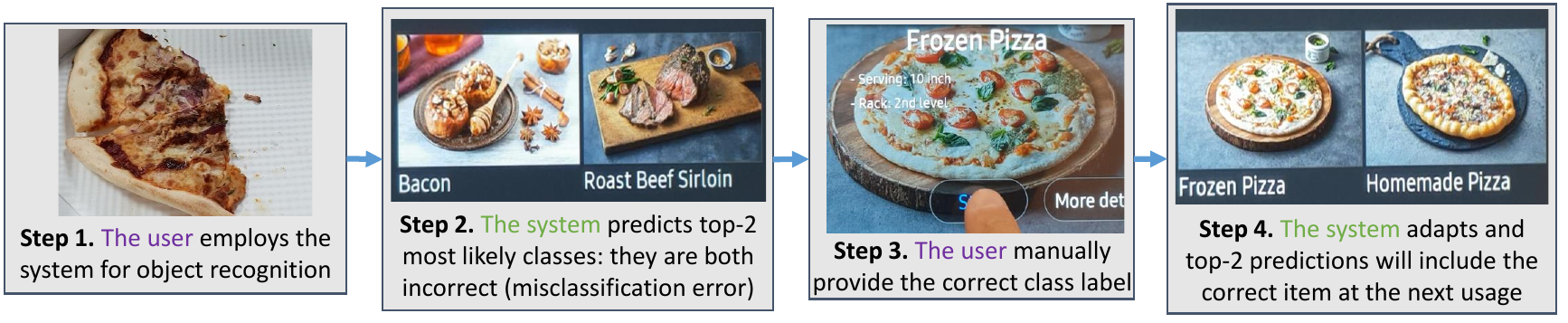}
  \caption{Proposed continual few-shot error correction system and user interactions flow in a food recognition use case.}
  \label{fig:teaser}
\end{teaserfigure}

\received{20 February 2007}
\received[revised]{12 March 2009}
\received[accepted]{5 June 2009}

\maketitle

\section{Introduction}
The integration of AI functionalities into everyday digital devices has grown exponentially, particularly in discriminative applications like image classification and object detection \cite{numalis}. From smart home appliances to mobile devices, AI models are becoming increasingly prevalent in enhancing user experience through automated recognition, classification, and detection tasks. However, these AI models frequently make prediction errors that frustrate users and diminish product reliability \cite{comprehensive}. This is particularly problematic in consumer devices where users expect consistent, accurate performance and have no means to correct persistent errors.

The challenge of AI model errors is compounded by several factors in real-world applications. First, the deployment environment often differs significantly from the training environment, leading to domain shift issues. Second, personal devices frequently encounter user-specific patterns and preferences that were not represented in the training data. Third, resource constraints on edge devices limit the applicability of traditional error correction approaches that require substantial computational power or storage.

Current approaches to handling AI errors primarily focus on detection rather than correction \cite{qiu2022detecting,hendrycks2016baseline,jiang2018trust}. When correction methods exist, they typically require substantial resources: either collecting additional training data \cite{teblunthuis2024misclassification}, finetuning models \cite{nartker2023will}, or maintaining large storage capacities \cite{corbiere2019addressing}. Such requirements make these solutions impractical for consumer devices with limited computational and storage resources. Moreover, these approaches often necessitate cloud connectivity for model updates, raising privacy concerns and limiting functionality in offline scenarios.

Furthermore, existing error correction methods often suffer from catastrophic forgetting, where correcting one type of error leads to degraded performance on previously well-handled cases. This creates a frustrating user experience where fixing one issue simply leads to another, ultimately reducing trust in the AI system. The challenge lies in developing a system that can adapt to user corrections while maintaining overall model performance and operating within device constraints.

We present a novel system that enables efficient, on-device error correction for AI models. Our solution allows users to correct model mistakes using just a few labelled examples without requiring significant computational resources or storage. For instance, as shown in Figure~\ref{fig:teaser}, users can correct misclassification errors with just one example (1-shot error correction), making the system highly practical for real-world applications. 

Our technical approach combines three key innovations that address the challenges of on-device error correction:
\begin{itemize}
\item Server-side knowledge distillation from foundation models to create robust yet compact models suitable for resource-constrained devices, leveraging the power of large models while maintaining practical deployability;
\item A lightweight, prototype-based classification pipeline provided with an efficient prototype update mechanism that allows continual adaptation to user corrections while preventing catastrophic forgetting;
\item A practical demonstration via an Android application that showcases real-time error correction capabilities, validating our approach in practical scenarios.
\end{itemize}

Our system addresses several key requirements for actual deployment: 
(i) \textit{resource efficiency}, as the system operates within the computational and storage constraints of typical consumer devices;
(ii) \textit{privacy preservation}, as all error corrections happen locally on the device, ensuring user data remains private;
(iii) \textit{ease of use}, as users can correct errors through simple interactions;
(iv) \textit{stability}, as the system maintains overall performance while adapting to user corrections;
(v) \textit{offline operation}, as the system does not require cloud connectivity.

Experimental results demonstrate the effectiveness of our system, achieving over 50\% error correction accuracy in one-shot scenarios while maintaining minimal computational overhead. We validate these results through both benchmark evaluations and real-world testing via our Android demonstration app. Our experiments cover a range of scenarios, from standard image classification tasks to more complex object detection applications, showing the system's versatility across different use cases.

The rest of this paper is organized as follows: Section~\ref{sec:related} discusses related work and existing approaches to AI error correction.
Section~\ref{sec:problem} formulates the problem setup and introduces our evaluation metrics. 
Section~\ref{sec:system} provides a detailed description of our system architecture, including both server-side and device-side components.
Section~\ref{sec:results} presents our experimental results and demonstration implementation. 
Finally, Section~\ref{sec:conclusion} concludes the paper.

\section{Related Work}
\label{sec:related}
Existing work related to AI model errors can be categorized into three main approaches.

\textit{1) Error Detection and Confidence Estimation:} most current solutions focus solely on analysing \cite{sikar2024misclassification} or detecting misclassifications \cite{qiu2022detecting,hendrycks2016baseline,jiang2018trust,corbiere2019addressing,nartker2023will}. 
While these methods are effective at identifying errors, they don't provide correction mechanisms. 
In \cite{manai2024minimizing} misclassifications of trained models are reduced thanks to the addition of a regularization term in the optimization process. 
Related approaches \cite{qu2023towards,qu2022improving} estimate model confidence to implement quality of service policies, such as discarding low-confidence predictions. In contrast, our work assumes users can provide ground truth labels for misclassified samples and focuses on actually correcting these errors with no need for backpropagation.

\textit{2) Domain Generalization and Adaptation:} several approaches attempt to improve model robustness through domain generalization at pre-training, aiming to handle multiple domains or adversarial samples \cite{barbato2024acmmmsys,chen2021amplitude,hendrycks2019augmix,zhou2022domain}. Similarly, domain adaptation techniques \cite{ben2006analysis,you2019universal,toldo2020unsupervised,ganin2015unsupervised} focus on adapting networks to domains different from their pre-training distribution. However, these approaches typically require large amounts of labelled data and don't address the correction of individual model mistakes.

\textit{3) Distribution Shift Detection:} a well-established research direction focuses on out-of-distribution detection, aiming to identify when input samples come from distributions unseen during pre-training \cite{yang2024generalized,camuffo2024enhanced,liu2020energy,liang2017enhancing}. While these methods can identify potentially problematic inputs, they differ fundamentally from our approach as they do not provide mechanisms to adapt models and correct previous misclassifications.

Our work provides a practical and resource-efficient solution for error correction that operates directly on user devices. Unlike previous methods, our system enables immediate error correction with minimal user input while maintaining low computational and storage requirements. 
By focusing on prototype updates rather than model retraining, we achieve efficient personalization without the need for extensive data collection or computational resources.

\section{Problem Formulation}
\label{sec:problem}
We address the challenging problem of correcting misclassification errors in AI models deployed on resource-constrained devices. While this is a critical issue for practical AI applications, it has received limited attention in existing literature. Our focus is on discriminative AI models for computer vision applications.

\subsection{Problem Setup}
\label{sec:problem:setup}
Our system's deployment process consists of three main phases.
\paragraph{Device Requirements Analysis}
The process begins with analyzing the target device's capabilities and constraints. We consider:
\begin{itemize}
\item hardware constraints (memory, compute, power, storage);
\item maximum allowable inference time;
\item minimum required accuracy.
\end{itemize}

Based on these constraints, we select an appropriate model architecture $M_I$ that offers the optimal accuracy-footprint tradeoff. Additional optimizations, such as quantization to 4-bit precision, may be applied if supported by the target hardware.

\paragraph{Server-side Pre-training}
The selected model $M_I$ undergoes pre-training on a training dataset $\mathcal{D}_{train}$ to produce the trained model $M_T$. The training samples in the dataset are indexed by $i=1,\dots,n$ and are composed of input images $\mathbf{X}_i$ and corresponding ground truth label $y_i \in \mathcal{C}$, namely $\mathcal{D}_{train}=\{ (\mathbf{X}_i,y_i) \}_{i=1}^n$.
The pre-training techniques employed in our system are described in Section~\ref{sec:system:serverside}.

\paragraph{On-device Error Correction}
After deployment, the model $M_T$ processes a stream of data $\mathcal{D}_{test}={ (\mathbf{X}_i^{test}, y_i^{test}) }_{i=1}^{n_{test}}$, where labels $y_i^{test}$ are only available when provided by users. 
The test dataset naturally partitions into: (i) correctly classified samples $\mathcal{D}_C\subset \mathcal{D}_{test}$, and (ii) misclassified ones $\mathcal{D}_E\subset \mathcal{D}_{test}$, where $\mathcal{D}_E \cap \mathcal{D}_C = \emptyset$.

Users can provide ground truth labels for $s$ samples from $\mathcal{D}_E$ for some classes, enabling few-shot error correction. The model $M_T$ is then adapted using these annotations to produce $M_A$, with improved performance on previously misclassified samples. The adaptation is efficient and backpropagation-free (see Section~\ref{sec:system:deviceside}).

\subsection{Evaluation Metrics}
We evaluate our system via three key metrics. We refer to \textit{accuracy} as the percentage of correctly classified images out of the total number of images in the set. 
We computed:
\begin{enumerate}
    \item \textbf{Base Recognition Accuracy ($\uparrow$):} Accuracy of $M_T$ on the test set $\mathcal{D}_{test}$, named $Acc_{base}$
    \item \textbf{Error Correction Accuracy ($\uparrow$):} Accuracy of $M_A$ on the misclassified set $\mathcal{D}_{E}$, named $Acc_E$
    \item \textbf{Forgetting Rate Percentage ($\downarrow$):} we compute the accuracy of the adapted model $M_A$ on the previously correctly-classified samples $\mathcal{D}_C$, named $Acc_C$. Then, the forgetting rate is computed as: $For := 100 - Acc_C$.
\end{enumerate}

Therefore, a successful error correction system should achieve: 
(i)  high error correction accuracy ($Acc_E$), 
(ii) minimal user labelling requirement (small $s$ for low-shot learning), 
(iii) low forgetting rate ($For$),
(iv) controlled footprints.

Our solution ensures deterministic control over inference time and memory usage increases, with built-in mechanisms to handle resource constraints (detailed in Section~\ref{sec:system:deviceside}).

\section{Our System}
\label{sec:system}

Our system consists of two sequential pipelines: (1) server-side pre-training followed by (2) on-device deployment and error correction. We first present the system for image classification and then extend it to object detection in Section~\ref{sec:system:objectdetection}.

\begin{figure*}
    \centering
    \includegraphics[width=0.9\linewidth, trim=0cm 8.3cm 3cm 0cm, clip]{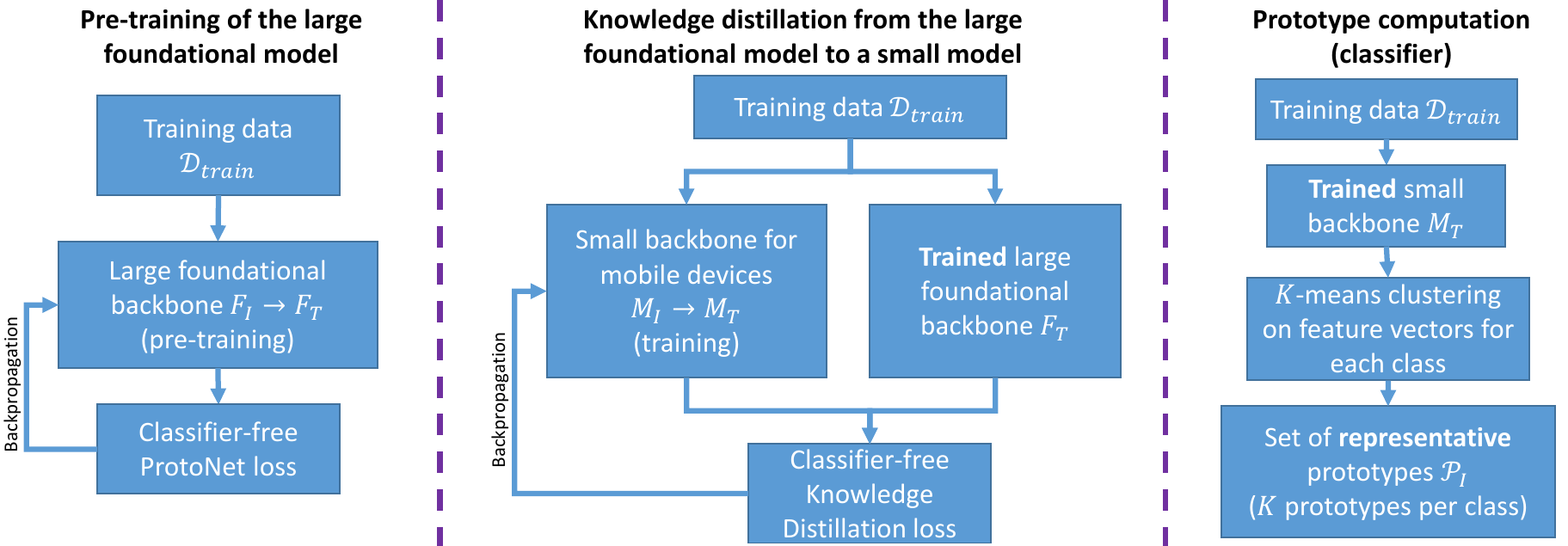}
    \caption{The server-side image classification pipeline is composed of three stages: (i) training of a large foundation model (Section~\ref{sec:system:serverside:foundation}, (ii) knowledge distillation from the trained foundation model to a smaller model (Section~\ref{sec:system:serverside:distillation}), and (iii) computation of prototypes as the classifier (Section~\ref{sec:system:serverside:prototypes}).}
    \label{fig:serverside}
\end{figure*}

\subsection{Server-side Pipeline}
\label{sec:system:serverside}

The server-side pipeline consists of three main stages, as illustrated in Figure~\ref{fig:serverside}.

\subsubsection{Foundation Model Pre-training}
\label{sec:system:serverside:foundation}

Motivated by the recent success of large foundation vision models, we leverage the transformer-based DINO-v2 \cite{oquab2023dinov2} model, finetuning it on domain-specific training data $\mathcal{D}_{train}$ at the server side (e.g., food data for food recognition applications). 
More specifically, starting from pre-trained weights ($F_I$), we train DINO-v2 without an output classifier (i.e., no output linear layer) using ProtoNet loss \cite{snell2017prototypical} to obtain $F_T$.
Intuitively, the ProtoNet loss learns a metric space where classification can be performed by computing distances to class prototypes, providing a robust feature extractor while maintaining flexibility compared to a linear classifier head.

\subsubsection{Knowledge Distillation into Small Model}
\label{sec:system:serverside:distillation}

Due to the limited-resource constraints on the device side, the finetuned large foundation model ($F_T$) may not satisfy the target footprints (e.g., model size and inference time) on the device (see Section~\ref{sec:problem:setup}). 
If this is the case, we select a smaller model $M_I$ (e.g., the convolutional-based MobileNet-V2 \cite{sandler2018mobilenetv2}). We load the pre-trained weights and continue training with knowledge distillation \cite{hinton2015distilling} from the frozen large (teacher) model $F_T$ over the training data $\mathcal{D}_{train}$, obtaining $M_T$.
As the distillation loss, we use an L1 distance between the class tokens of the teacher and student models, namely:
\begin{equation}
    \mathcal{L}_{KD} = \frac{1}{n} \sum_{i=1}^n | F_T(\mathbf{X}_i) - M(\mathbf{X}_i) |,
\end{equation}
where $M$ is initialized as $M_I$ and trained to obtain $M_T$.

\subsubsection{Prototype Computation}
\label{sec:system:serverside:prototypes}
So far, we considered models without any classifier head. Therefore, the final stage at the server side is to compute a set of representative prototypes for each class $c$ in $\mathcal{D}_{train}$ using the trained model $M_T$.

Specifically, we compute the feature vectors of every sample, $\mathbf{v}_{\mathbf{X}_i} = M_T(\mathbf{X}_i), \forall i=1,\dots,n$, and compute a $K$-means clustering for each class $c$. Namely, defining as $\mathcal{V}$ the set of features associated to ground truth label $c$, i.e., $\mathcal{V}={\{\mathbf{v}_{\mathbf{X}_i}\}_{\forall i, \mathrm{s.t.} y_i=c}}$, the objective of the $K$-means for class $c$ is to find:
\begin{equation}
    \arg\min_\mathcal{V}
    \sum_{j=1}^K \sum_{\mathbf{v}\in \mathcal{V}} || \mathbf{v} - \mathbf{p}_{c,j}  ||^2
\end{equation}
where the $K$ cluster centroids are denoted as $\mathcal{P}_c=\{\mathbf{p}_{c,j}\}_{j=1}^K$ and are the representative prototypes for class $c$. In other words, $\mathcal{P}_c$ is a set containing the $K$ representative prototypes for class $c$.

The set of all representative prototypes across all classes together with their corresponding class labels is denoted as $\mathcal{P}_I=\{(\mathcal{P}_c, c)\}, \forall c\in \mathcal{C}$.

\begin{figure}
    \centering
    \includegraphics[width=\linewidth, trim=0cm 7.3cm 13.9cm 0cm, clip]{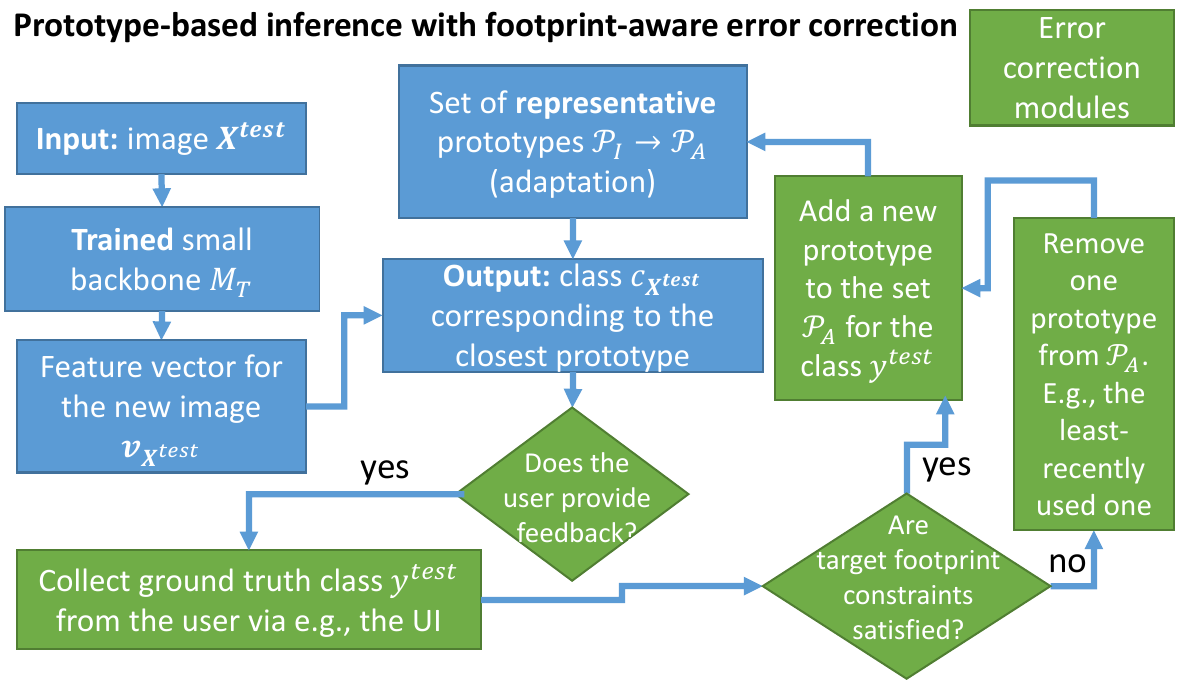}
    \caption{The device-side image classification pipeline with error correction blocks in green.}
    \label{fig:deviceside}
\end{figure}

\subsection{Device-side Pipeline}
\label{sec:system:deviceside}

The device-side pipeline is shown in Figure~\ref{fig:deviceside}.

\paragraph{Classification Process}
The system receives an input image $\mathbf{X}^{test}$.
The image is passed through the small backbone $M_T$ (pre-trained on the server) to obtain the feature vector ${\mathbf{v}_{\mathbf{X}^{test}} = M_T(\mathbf{X}^{test})}$.
We initialize a new set of adapted prototypes $\mathcal{P}_A=\mathcal{P}_I$. 
Then, we compute the cosine distance between the feature vector $\mathbf{v}_{\mathbf{X}^{test}}$ and all prototypes ${\textbf{p}}_{c,j}\in \mathcal{P}_c, \forall c \in \mathcal{C}, \forall j = 1, \dots, K$ from the set of representative prototypes $\mathcal{P}_A$ (constructed on the server). Finally, we select the class corresponding to the minimum cosine distance as the predicted inferred class $c_{\textbf{X}^{test}}$. 
In other words:
\begin{equation}
\label{eq:classification}
    c_{\textbf{X}^{test}} := \arg \min_c (\cos\mathrm{dist}(\mathbf{v}_{\mathbf{X}^{test}}, \mathcal{P}_c)).
\end{equation}
We note that cosine distance function $\cos\mathrm{dist}()$ computes the cosine distance between $\mathbf{v}_{\mathbf{X}^{test}}$ and all elements of $\mathcal{P}_c$.

\paragraph{Error Correction}

If the inferred class $c_{X^{test}}$ is incorrect (i.e., the prediction is different from the ground truth $y^{test}$), the user could optionally decide to manually annotate the image $\mathbf{X}^{test}$ with the correct label $y^{test}$, via e.g., the user interface (UI).
Then, our error correction system updates the array of representative prototypes: 
\begin{align}
    \mathcal{P}_A := &\{ (\mathcal{P}_c, c) \in \mathcal{P}_A, \ \ c\neq y^{test} \} 
    \ \ \ \cup\\
    & \{ (\mathrm{add}(\mathcal{P}_{y^{test}}, \mathbf{v}_{\mathbf{X}^{test}}), y^{test}),\ \ \  (\mathcal{P}_{y^{test}}, y^{test}) \in \mathcal{P}_A \}.
\end{align}
The first term ensures that representative prototypes of classes other than the current one are not modified.
The second term adds the current feature vector $\mathbf{v}_{\mathbf{X}^{test}}$ to the current list of representative prototypes for class $y^{test}$, i.e., $\mathcal{P}_{y^{test}}$.

To maintain resource constraints, the system drops one prototype from the set $\mathcal{P}_A$ whenever the maximum memory or footprint target limits are hit. For example, the least-recently-used prototype can be removed.
The system continually updates $\mathcal{P}_A$ based on user feedback by adding incoming user-annotated feature vectors and uses $\mathcal{P}_A$ to predict the class of any follow-up images.

\subsection{Extension to Object Detection}
\label{sec:system:objectdetection}

Extending the system outlined so far to object detection can be done easily by incorporating an object detection network and image slicing mechanism based on the predicted bounding boxes.

\paragraph{Server-side Modifications.} On the server side, we incorporate a state-of-the-art efficient object detection network (e.g., YoloV8 \cite{yolov8_ultralytics}) and finetune it to predict bounding boxes of objects of interest. In the food domain example, the object detector could be trained to identify bounding boxes corresponding to food items in the scene (i.e., one-class classification).

\paragraph{Device-side Implementation.} On the device side, we use the finetuned model to extract the bounding boxes of the test image. Then, for each bounding box, we crop the image region and process it through the image classification network. 

\paragraph{Implementation Trade-offs}
Our current implementation uses separate networks for detection and classification, which offers maximum accuracy but higher computational cost. 

To save complexity, one could use detection backbone features directly for classification. 
This optimization is demonstrated in \cite{barbato2024cross}. 
The single backbone solution would reduce computational overhead at a cost of less than 5\% relative accuracy.
While further optimizations are possible by building prototypical inference directly on detection features, we leave this exploration for future work to maintain focus on our core contribution of error correction.

\section{Experimental Results}
\label{sec:results}

\subsection{Experimental Setup}

\subsubsection{Implementation Details} 
The server-side pipeline is implemented in Python using PyTorch, and training is conducted on a single NVIDIA GeForce RTX 3090 Ti GPU. Model quantization and initial server-side testing of the quantized models were implemented on an AMD Ryzen 9 5950X CPU. For on-device demonstrations, the model $M_T$ is exported to ONNX and then converted to TFLite, followed by uniform 8-bit quantization. 
The quantized TFLite model is used with the TFLite C++ interpreter for inference. The device-side pipeline is integrated into NNTrainer \cite{moon2024new} for deployment on Android devices. Demonstrations are conducted on a Samsung Galaxy S24 Ultra smartphone.

The complete codebase for both server-side and device-side pipelines will be open-sourced upon acceptance.

\subsubsection{Models} 
For the Image Classification task, the transformer-based foundation model DINO-V2-small \cite{oquab2023dinov2} is distilled into an efficient convolutional MobileNet-V2 \cite{sandler2018mobilenetv2}.
For the Object Detection task, the YoloV8-nano model \cite{yolov8_ultralytics} is utilized.

\subsubsection{Hyperparameters}

For the server-side pre-training, we train DINO-v2-small for 100 epochs with a batch size of 256 and a learning rate of 1e-5.
Knowledge distillation onto MobileNet-V2 is done for 100 epochs with a batch size of 256 and a learning rate of 0.1.
Finally, we used YoloV8-nano \cite{yolov8_ultralytics} as the object detection model and we finetuned it on the OpenImages-V7 \cite{kuznetsova2020open} dataset reaching 0.347 mAP\@50.
We trained it for 200 epochs with early stopping enabled with patience 20 epochs, using a batch size of 16 and a learning rate of 0.01.
For the server-side prototypes computation, we set $K=3$, i.e., we store the three most representative prototypes per class.

\subsubsection{Datasets}
We used two datasets: the Food-101 \cite{bossard2014food} and the Oxford Flowers-102 \cite{nilsback2008automated}, containing 101 and 102 classes, respectively. 
Both datasets represent domain-specific applications (e.g., food or flower recognition), where the user may want to provide the correct label to improve the AI model. 
As these datasets are out-of-domain for DINO-V2, domain finetuning is required. Each dataset is split into 70\% training, 15\% validation, and 15\% testing per class; i.e., we preserve 70\% of the images of each class in the training set.
Few-shot error correction experiments are conducted with $s \in \{1, 2, 3, 4, 5, 7, 10, 20, 50\}$ annotated samples per class.

\begin{table}[t]
  \centering
  \setlength{\tabcolsep}{9pt}
  \caption{Error correction accuracy and forgetting rate of our system on two image classification datasets.}
    \begin{tabular}{cccrr}
    \toprule
          & \multicolumn{2}{c}{\textbf{Food-101}} & \multicolumn{2}{c}{\textbf{Flowers-102}} \\ \cmidrule(lr){2-3} \cmidrule(lr){4-5}
    \textbf{\# shots} $s$ & $Acc_E \uparrow$ & $For \downarrow$ & \multicolumn{1}{c}{$Acc_E \uparrow$} & \multicolumn{1}{c}{$For \downarrow$} \\\midrule
    \textbf{1}  & 51.1\% & 0.018\% & 54.3\% & 0.011\% \\
    \textbf{2}  & 67.3\% & 0.053\% & 68.9\% & 0.038\% \\
    \textbf{3}  & 75.0\% & 0.118\% & 78.4\% & 0.094\% \\
    \textbf{4}  & 79.9\% & 0.157\% & 81.1\% & 0.114\% \\
    \textbf{5}  & 83.2\% & 0.179\% & 85.7\% & 0.162\% \\
    \textbf{7}  & 87.3\% & 0.202\% & 89.0\% & 0.179\% \\
    \textbf{10} & 91.0\% & 0.237\% & 92.6\% & 0.195\% \\
    \textbf{20} & 93.2\% & 0.447\% & 96.7\% & 0.364\% \\
    \textbf{50} & 96.4\% & 0.870\% & 98.8\% & 0.577\% \\
    \bottomrule
    \end{tabular}%
  \label{tab:main_results}%
\end{table}%

\subsection{Image Classification Results}

Quantitative results of our system are reported in Table~\ref{tab:main_results} in terms of error correction accuracy $Acc_E$ and forgetting rate $For$ for the image classification task. 
The base recognition accuracy $Acc_{base}$ of the $M_T$ model with initial representation prototypes $P_I$ is 90.6\% and 94.3\% on the Food-101 and the Flowers-102 datasets, respectively.

Our system allows user to correct most of the misclassification errors with limited feedback.
For example, correcting 51.1\% of classification errors on Food-101 requires the user to annotate just one sample per class. 
At the same time, only 0.018\% of previously-corrected samples are now misclassified by the adapted model.

\subsubsection{Ablation Studies}

\paragraph{Prototypes vs.\ Linear Classifier Head.} First, we checked that using a fully connected linear classifier head does not produce higher base classification accuracy ($91.1\%$ on Food-101 and $93.6\%$ on Flowers-102) than prototypical inference ($90.6\%$ and $94.3\%$, respectively). 
Furthermore, fully-connected layers are less adaptable, and mistakes are harder to correct with efficient updates.

\begin{table*}[t]
  \centering
  \setlength{\tabcolsep}{9pt}
  \caption{Ablation studies on the server-side pipeline on the Food-101 dataset. $Acc_{base}$ for all approaches is (from left to right): $82.6$, $76.4$, $92.4$, and $90.6\%$.}
    \begin{tabular}{ccccccccc}
    \toprule
          & \multicolumn{2}{c}{$M_T$ only} & \multicolumn{2}{c}{$F_I$} & \multicolumn{2}{c}{$F_T$} & \multicolumn{2}{c}{Ours} \\ \cmidrule(lr){2-3} \cmidrule(lr){4-5} \cmidrule(lr){6-7} \cmidrule(lr){8-9}
    \textbf{\# shots} $s$ & $Acc_E \uparrow$ & $For \downarrow$ & $Acc_E \uparrow$ & $For \downarrow$ & $Acc_E \uparrow$ & $For \downarrow$ & $Acc_E \uparrow$ & $For \downarrow$  \\\midrule
    \textbf{1}  & 0.6\% & 0.1\% 
    & 48.7\% & 0.1\%
    & 15.1\% & 0.1\%
    & 51.1\% & 0.0\%
    \\
    
    \textbf{3}  & 7.0\% & 0.1\% 
    & 72.3\% & 0.3\%
    & 48.3\% & 0.1\%
    & 75.0\% & 0.1\% \\
    
    \textbf{5}  & 13.2\% & 0.3\%
    & 82.9\% & 0.4\%
    & 57.3\% & 0.1\%
    & 83.2\% & 0.2\% \\
    
    \textbf{10} & 21.7\% & 0.6\% 
    & 91.2\% & 0.6\%
    & 65.4\% & 0.2\%
    & 91.0\% & 0.2\% \\
    
    \bottomrule
    \end{tabular}%
  \label{tab:ablation}%
\end{table*}%

\begin{figure*}
    \centering
    \includegraphics[width=0.75\linewidth, trim=0cm 0cm 0cm 1.8cm, clip]{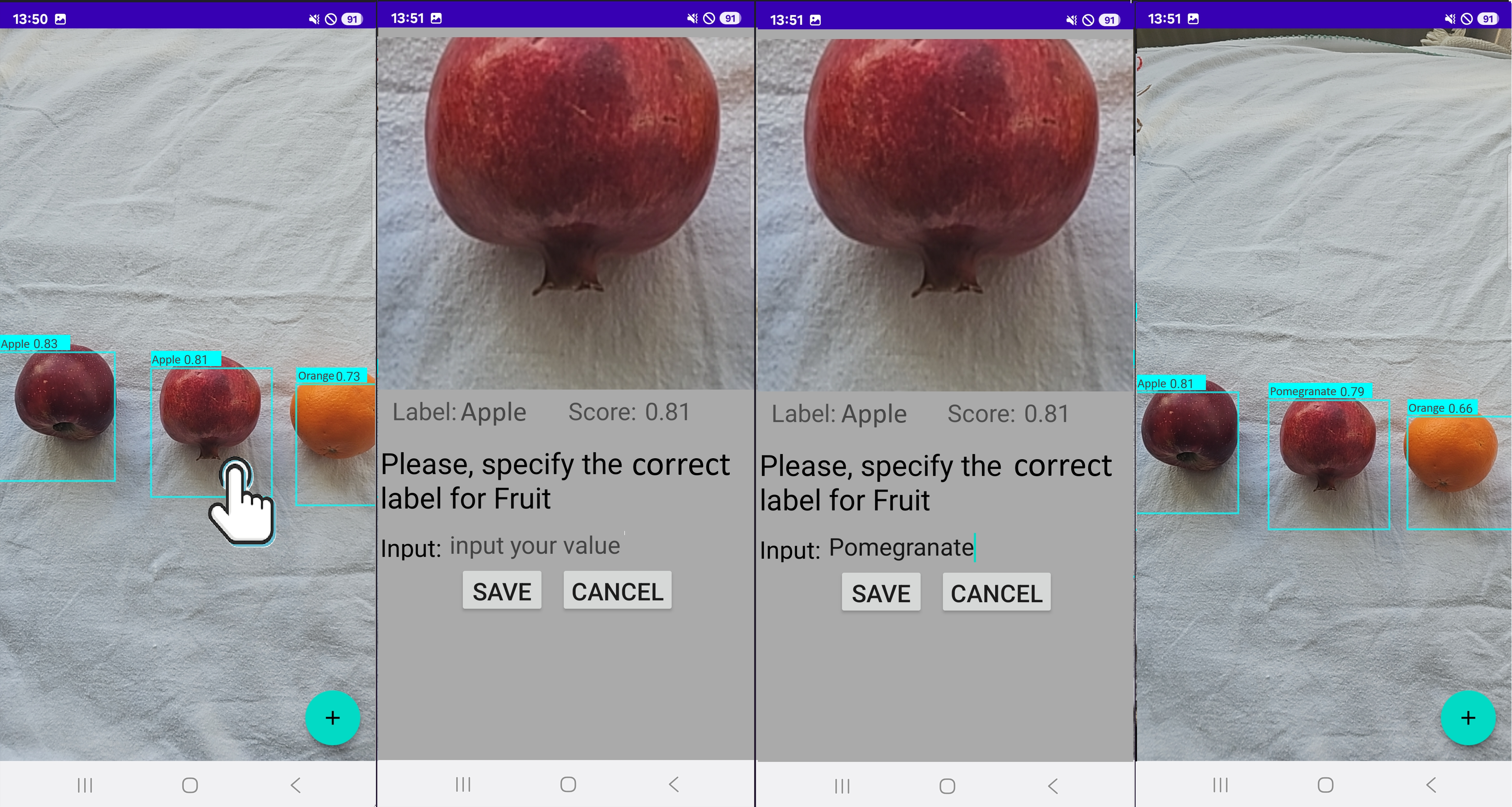}
    \caption{Application workflow of our demo: (1) the user identifies misclassified objects and clicks anywhere within their bounding box, (2-3) the user provides the correct label, and (4) the user observes improved classification in subsequent iterations.}
    \label{fig:ondevice_results}
\end{figure*}

\paragraph{Server-side Pipeline.} 
Table~\ref{tab:ablation} highlights the performance of different server-side pre-training strategies on Food-101.
Using only the small model finetuned on domain-specific data, i.e., $M_T$, results in low $Acc_E$ due to limited feature richness.
Using the pre-trained foundation model $F_I$ gives a good balance between error correction accuracy and forgetting, nonetheless the performance on the base classes is quite poor as the model is pre-trained on a large generic dataset and not domain-finetuned. Additionally, the foundation model is typically an order of magnitude larger than the selected small model.
The finetuned foundation model $F_T$ overfits to the domain-specific data, achieving high $Acc_{base}$ but poor adaptability to correct mistakes efficiently.
Finally, our approach strikes an optimal balance, sacrificing only a small amount of error recognition accuracy (2.2\%) to achieve significantly improved error correction capability and minimal forgetting, all while being more computationally efficient than foundation model-based methods.

\subsection{On-device Demonstration}

In our demo, we focus on the object detection task due to its practical value where users point a camera to target objects.
Both MobileNet-V2 and YoloV8-nano are quantized to 8 bits, achieving a $\sim3\times$ inference speedup with negligible accuracy degradation (<1\%) compared to the 32-bit models. 
For instance, on a Samsung Galaxy S24, YoloV8-nano achieves 40 ms inference time (compared to 110 ms for 32-bit) while maintaining a mAP@50 of 0.347 (from 0.350). The 8-bit model has a compact size of 3.2 MB, enabling real-time inference on consumer devices.

Overall, the system runs in real-time on consumer smartphones.
Figure~\ref{fig:ondevice_results} illustrates the application workflow. 
In the first screenshot, the user can benefit from the detection model to identify some food items. In case of a misclassification error, the user may decide to provide the correct ground truth: in our app, the user should click on the bounding box and enter the correct label (second and third screenshots). Finally, the system updates its internal prototypical representations and will correctly identify the item at subsequent iterations (third screenshot).

\section{Conclusion}
\label{sec:conclusion}

In this work, we introduced a simple yet effective system for few-shot continual error correction in discriminative AI models, such as those used for object recognition. Our approach empowers end users to optionally and seamlessly correct misclassification errors in AI models, enhancing their overall utility and reliability.
The proposed system comprises a server-side pre-training stage to develop a robust feature extractor while adhering to the resource constraints of target devices, such as memory, storage, and inference time. Additionally, we designed a novel on-device mechanism for few-shot continual error correction, which efficiently updates the stored prototypical representations.
Our system represents a step forward in enabling pervasive AI applications, setting new benchmarks for accuracy, adaptability, and practicality in consumer products.

\bibliographystyle{ACM-Reference-Format}
\bibliography{refs}

\end{document}